\documentclass[sigconf]{acmart}
\AtBeginDocument{%
  }

\setcopyright{acmlicensed}
\copyrightyear{2024}
\acmYear{2024}
\acmDOI{}
\acmConference[CSAI '24]{2024 8th International Conference on Computer Science and Artificial Intelligence}{December 06--08,
  2024}{Beijing, China}
\acmISBN{}

\usepackage{caption}

\begin{document}

\title{VistaFlow: Photorealistic Volumetric Reconstruction with Dynamic Resolution Management via Q-Learning}

\author{Jayram Palamadai}
\email{jpalamadai@imsa.edu}
\orcid{0009-0004-4882-9595}
\affiliation{%
  \institution{Illinois Mathematics and Science Academy}
  \city{Aurora}
  \state{Illinois}
  \country{USA}
}

\author{William Yu}
\email{wyu@imsa.edu}
\affiliation{%
  \institution{Illinois Mathematics and Science Academy}
  \city{Aurora}
  \state{Illinois}
  \country{USA}
}

\renewcommand{\shortauthors}{Palamadai et al.}

\begin{abstract}
  We introduce VistaFlow, a scalable three-dimensional imaging technique capable of reconstructing fully interactive 3D volumetric images from a set of 2D photographs. Our model synthesizes novel viewpoints through a differentiable rendering system capable of dynamic resolution management on photorealistic 3D scenes. We achieve this through the introduction of QuiQ, a novel intermediate video controller trained through Q-learning to maintain a consistently high framerate by adjusting render resolution with millisecond precision. Notably, VistaFlow runs natively on integrated CPU graphics, making it viable for mobile and entry-level devices while still delivering high-performance rendering. VistaFlow bypasses Neural Radiance Fields (NeRFs), using the PlenOctree data structure to render complex light interactions such as reflection and subsurface scattering with minimal hardware requirements. Our model is capable of outperforming state-of-the-art methods with novel view synthesis at a resolution of 1080p at over 100 frames per second on consumer hardware. By tailoring render quality to the capabilities of each device, VistaFlow has the potential to improve the efficiency and accessibility of photorealistic 3D scene rendering across a wide spectrum of hardware, from high-end workstations to inexpensive microcontrollers.
\end{abstract}

\ccsdesc[500]{Computing methodologies~Volumetric models}
\ccsdesc[500]{Computing methodologies~Rasterization}

\keywords{Three-dimensional (3D) rendering, Volumetric reconstruction, Differentiable rendering, Dynamic resolution rendering, Q-learning}
\begin{teaserfigure}
  \includegraphics[width=\textwidth]{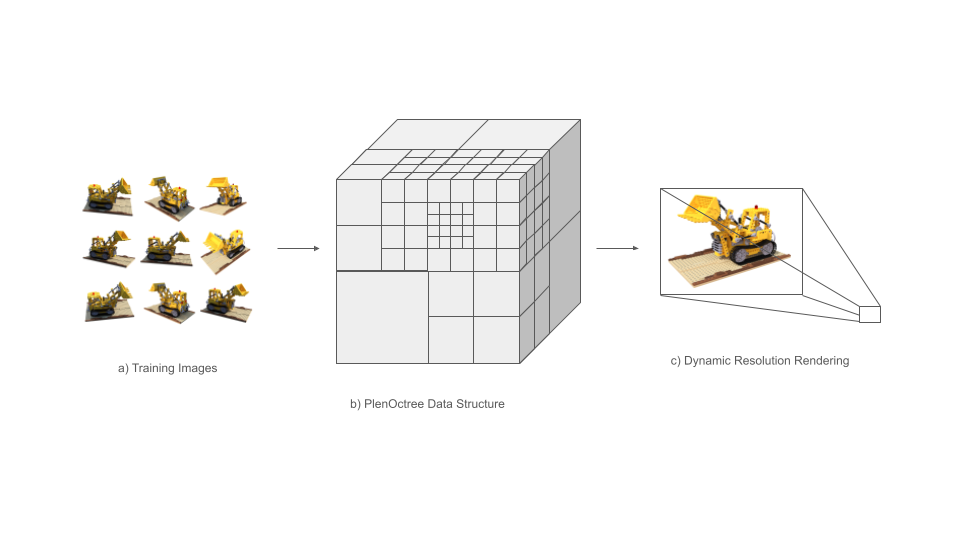}
  \caption{Overview of the VistaFlow optimization model. We begin with (a) a set of input images, which are then (b) converted into the PlenOctree dataset via direct optimization. This is then used to train the QuiQ dynamic resolution controller, which is described in Fig 2.
}
\end{teaserfigure}

\received{5 February 2025}

\maketitle

\section{Introduction}
The recent advent of artificial intelligence has drastically expanded the scope of 3D image generation in the past few years. What once needed to be meticulously modeled by hand can now be created by a program using only a few pictures taken with a standard camera \cite{Yu21plenoctrees}. Although the existing code for this process excels at reducing model generation time, the employed processes are computationally taxing. Interaction with these generated models thus generally results in suboptimal frame rates when conducted with standard graphics cards. Current algorithms can process 2D image information into hierarchical data structures \cite{Seitz02}. This form of rendering reduces training times while maintaining high image quality compared to competing rendering methods.

Another significant factor contributing to computational load is the real-time rendering of complex scene interactions such as rotation. The calculations necessary for these operattions requires a tradeoff between image quality and framerate in order to produce a digestible result. QuiQ is an autonomous policy network trained to manage real-time operation of this quality-framerate tradeoff.

To summarize, we make the following constributions:
\begin{itemize}
\item VistaFlow, the first method that allows dynamic frame rate stabilization for radiance field rendering.
\item Highly efficient conversion of images to the PlenOctree data structure without reliance on GPU acceleration.
\item QuiQ, an autonomous policy network trained to selectively offset the computational load of complex scene interactions.
\item A scalable rendering framework that significantly lowers the barrier of entry for high-quality 3D graphics on consumer-grade hardware.
\end{itemize}

\section{Related Work}

\subsection{Classical Volume Reconstruction}

Classical Volume Reconstruction is a method used primarily in computerized tomography or medical imaging to construct a 3D image from a series of 2D cross-sectional images. This is often performed to visualize certain bodily structures. Within CT imaging, data is acquired by projecting X-rays through the body from multiple different angles. Detectors measure the absorption of the X-rays as they pass through different tissues. The data collected from these projections is then used to generate a series of 2D cross-sectional images that represent the internal structures of the measured body from multiple different depths. 

Classical Volume Reconstruction utilizes filtered back projection (FBP) to construct a 3D image using the newly generated series. FBP applies a series of filters to the raw data and then back-projects this into the form of a 3D image. Back-projection is the process of tracing the intensity values of the X-rays measured by the detectors along their path through the object \cite{Piala21}. Each 3D pixel, known as a voxel, is then assigned an intensity value by summing the contributions of each projection that passes the voxel. Once the 3D volume is constructed, it may be visualized and manipulated using volume rendering techniques.

\subsection{Neural Volume Reconstruction}

Neural Volume Reconstruction leverages the power of deep learning techniques for better image quality, faster reconstruction times, and the ability to handle more complicated reconstructive tasks. Similar to Classical Volume Reconstruction, neural volume reconstruction begins by acquiring 2D slices of a 3D object using image modalities such as CT, MRI, or other tomographic techniques. Following data acquisition, large datasets are constructed which contain subsets of 2D slices and their corresponding high-quality 3D volumes. Prior to training, datasets may undergo preprocessing steps such as normalization, cropping, or resampling to ensure consistency and compatibility with the neural network architecture. Data augmentation techniques such as rotation, translation, scaling, or adding noise may also be applied to increase the diversity of the training data and improve the generalization capability of the model.

A deep neural network architecture is then designed and trained to perform the task of reconstructing these volumes using the 2D slices. Convolutional Neural Networks (CNNs) are commonly used due to their ability to learn complex patterns and relationships within the data. During training, the neural network learns to map the input 2D projection images to their corresponding 3D volumes. The network also adjusts its internal parameters (such as weights/biases) through an optimization process, which minimizes the difference between the predicted 3D volumes and the ground truth volumes from the training dataset \cite{Knapitsch17}. A ground truth volume here is the ideal or reference 3D volume that most accurately and completely represents our given object. During the training and evaluation of a reconstruction algorithm, reconstructed volumes are compared to the ground truth volume to assess their accuracy and fidelity. Metrics such as structural similarity index (SSI), mean squared error (MSE), or visual inspection may be used to quantify the similarity between the reconstructed and ground truth volumes.

Once training has been completed, the neural network may be deployed for inference, where it takes new 2D projection images as input and reconstructs an original 3D volume. Along with the advantage of improved image quality and higher potential speeds, the training of the neural network enables the handling of more challenging reconstruction tasks such as noisy images or inputs with incomplete data \cite{Mildenhall20}.

\subsection{Accelerating NeRF}

Neural radiance fields (NeRFs) \cite{Aliev19} represent a newer approach in computer graphics and computer vision for generating realistic 3D scenes. These models capture the volumetric properties of a scene and predict the color and intensity of light at any point in space. NeRFs are typically trained on pairs of 3D points and corresponding observed RGB colors from known viewpoints. 

During training, the neural network learns to approximate the underlying volumetric function that describes the scene. In the rendering process, NeRFs generate images by sampling rays from the camera's viewpoint and computing the radiance values along each ray. This enables the creation of photorealistic images with realistic lighting, shading, and surface details. However, NeRFs can be computationally demanding, requiring significant resources for training and rendering \cite{Reiser21}. Ongoing research aims to improve the scalability and efficiency of NeRFs to make them applicable to a wider range of scenes and applications \cite{Barron21, Garbin21, Kerbl23}.

One such method of improvement comes in the form of the PlenOctree \cite{Yu21plenoctrees}. This method introduces an octree-based 3D representation enabling real-time rendering of NeRFs supporting view-dependent effects such as mirroring. This is achieved through the use of closed-form spherical basis functions, enabling it to render arbitrary geometrical scenes without quality loss. Overall, this approach can render 800×800 images at over 150 FPS, which is more than 3000 times faster than conventional NeRFs. 

\section{Method}

Our framework allows for direct training of the PlenOctree data structure without prior dependence on neural rendering or conversion from a sparse voxel grid. Our framework as outlined in Fig. 1 constructs this representation from 2D images by minimizing rendering loss on selected rays during training. During an initial demonstration of the final model that doubles as a data collection stage, rendering parameters are adjusted according to a predefined sequence of inputs. Our differentiable volume renderer, QuiQ, is then trained on this data to dynamically adjust ray sampling parameters in real-time in order to maintain framerate stability. We demonstrate this process in Fig. 2. 

\subsection{Differential Volume Rendering}

Following the success of previous differentiable volume rendering techniques \cite{Yu21plenoxels}, VistaFlow calculates the values used during rasterization \cite{Chen22} through a process of ray casting. The output color (3) is calculated by combining samples collected along the ray according to (2). Equation (1) represents the quantity of light along ray \textit{r} that reaches sample \textit{i}:

\begin{equation}
\Xi_i=e^{-\sum_{j=1}^{i-1} \sigma_j \delta_j}
\end{equation}

\begin{equation}
\Gamma_i=1-e^{-\sigma_i \delta_i}
\end{equation}

\begin{equation}
\hat{C}(r)=\sum_{i=1}^N c_i \Xi_{i-1} \Gamma_i\left(\Gamma_{i-1}-1\right)
\end{equation}

Our differentiable volume rendering technique is crucial to the VistaFlow framework; by enabling the back-propagation of errors through the rendering process, differential volume rendering allows efficient generation of the PlenOctree data structure.

\subsection{Trilinear Interpolation}

Accurately interpolating opacity values across a 3D grid is crucial for rendering semi-transparent volumes, where light can penetrate and scatter within the material. Trilinear interpolation is employed to calculate the opacity at any point within a voxel grid based on the opacity values at the eight vertices of the enclosing voxel.

The interpolation process can be mathematically expressed as follows. Let $V_{i,j,k}$ represent the opacity value at the vertex with indices $i,j,k$ within the grid, and $x,y,z$ denote the normalized coordinates of the target point within the voxel, ranging from 0 to 1. The trilinearly interpolated opacity, , at this point is computed by first performing linear interpolations along the x-axis, followed by interpolations along the y-axis, and finally along the z-axis:

\begin{displaymath}
\sigma(x, y, z)=\sum_{i=0}^1 \sum_{j=0}^1 \sum_{k=0}^1 V_{i, j, k} x_i y_j z_k(1-x)^{1-i}(1-y)^{1-j}(1-z)^{1-k}
\end{displaymath}

Trilinear interpolation facilitates the efficient use of computational resources by allowing detailed properties to be stored at discrete points within our PlenOctree grid, with intermediate values accurately estimated through interpolation. This balance between accuracy and efficiency is vital in real-time applications, such as video games and interactive simulations, as well as in high-fidelity rendering tasks in film and scientific visualization.

\begin{figure*}[!t]
  \centering
  \includegraphics[width=\linewidth]{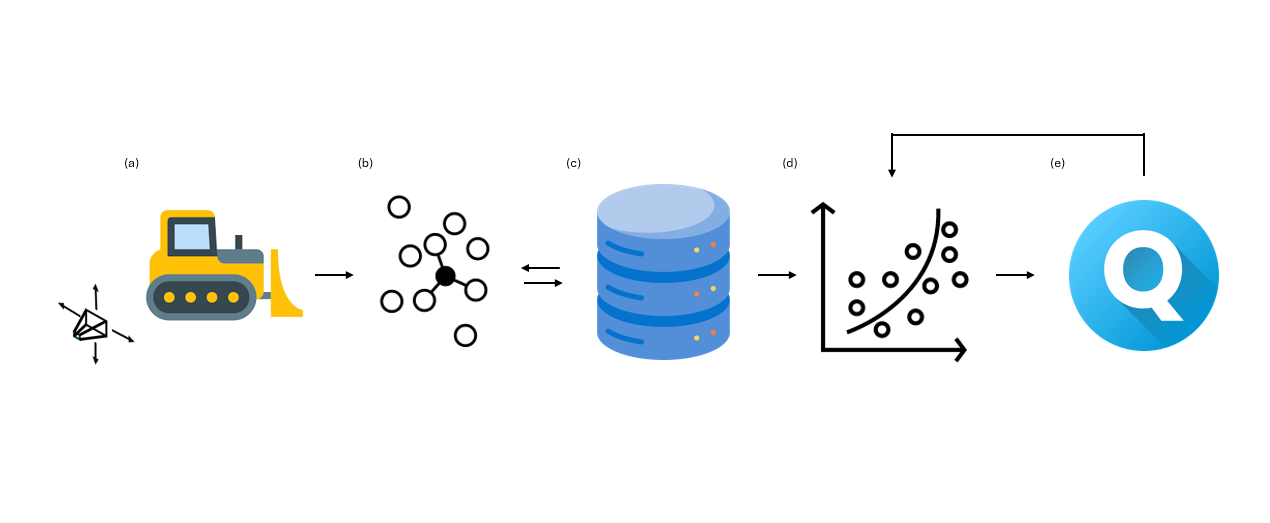}
  \caption{Overview of the QuiQ Training Process. We begin with a trained PlenOctree model as described in Fig 1. We then (a) benchmark the system by rendering the model along a predetermined set of camera movements. During this process, we collect framerate information to assess the effect of various rendering parameters on output quality. Next, we (b) use a k-Nearest Neighbors algorithm to find the most similar data from a (c) selection of prerecorded benchmark profiles. This allows us to effectively increase our dataset without the necessary collection time.  We use this increased dataset to (d)  train a reward function through ridge regression. Finally, this reward function is used to train the (e) dynamic resolution controller that ultimately controls the resolution parameters during output.
}
\end{figure*}

\subsection{PlenOctree Generation}

In the pursuit of achieving high-resolution imagery, our methodology employs a hierarchical refinement technique, commencing with a grid at a lower resolution. This process encompasses an initial phase of optimization, followed by the elimination of redundant voxels. The refinement continues with the subdivision of the remaining voxels, each bisected along every dimension, thereby facilitating further optimization. For instance, within synthetic environments, the procedure initiates with a grid of $256^3$ resolution, which is then enhanced to $512^3$. The trilinear interpolation technique is utilized for initializing grid values subsequent to each voxel subdivision, allowing for smooth transitions between different resolutions.

The voxel pruning process adheres to the framework established by PlenOctrees, which involves the application of a threshold to either the maximum weight or the density value of each voxel, evaluated across all training rays. However, the indiscriminate removal of voxels can inadvertently affect the fidelity of colors and densities near surfaces due to the interpolative nature of trilinear interpolation with adjacent exterior voxels. To circumvent this issue, a strategic dilation operation is employed, ensuring the preservation of essential details by pruning a voxel only if it and its immediate neighbors are concurrently considered superfluous.

Concurrently, our optimization strategy is geared towards refining voxel opacities and spherical harmonic coefficients to minimize the mean squared error (MSE) in rendered pixel colors, incorporating total variation (TV) regularization to enhance visual coherence. This optimization process leverages a stochastic sampling of rays and voxels for evaluating the MSE and TV terms, respectively, employing a tailored learning rate for both opacity and spherical harmonic adjustments, standardized across various scene types.

\begin{table*}[h!]
  \caption{PSNR compared to previous methods, by scene in the NeRF Synthetic dataset. Even split by scene, we outperform previous models in most scenes.
}
\begin{tabular}{lccccccc}
\toprule
        & \textbf{Chair} & \textbf{Drums} & \textbf{Ficus} & \textbf{Hotdog} & \textbf{Lego} & \textbf{Materials} & \textbf{Mic Ship Avg.} \\
\midrule
Gaussian Splatting     & 35.85          & 26.16          & 34.87          & 37.72          & 35.78         & 30.00              & 35.36                 \\
Point-NeRF             & 35.40          & 26.06          & 36.13          & 37.40          & 36.39         & 29.78              & 36.22                 \\
INGP-Base              & 35.00          & 25.48          & 36.81          & 34.10          & 29.14         & 33.26              & 31.83                 \\
Plenoxels              & 33.98          & 25.35          & 36.51          & 37.48          & 35.70         & 30.41              & 33.29                 \\
Mip-NeRF               & 35.14          & 26.02          & 36.02          & 37.32          & 36.02         & 29.72              & 33.79                 \\
\midrule
Ours                   & 30.80          & 33.32          & 36.39          & 36.02          & 36.51         & 30.52              & 34.22                 \\
\bottomrule
\end{tabular}
\end{table*}

\subsection{Dynamic Resolution Rendering}

As defined in Equation 3, the QuiQ dynamic resolution renderer controls $\delta_i$, the distance between samples, and $\gamma$, the threshold of light intensity below which a ray $r$ is terminated. In areas with high detail or rapid changes in color or intensity, $\delta_i$ is reduced to capture these nuances more accurately. Conversely, in regions with minimal variation, $\delta_i$ is increased, reducing the number of samples required and thus improving performance. This adaptive approach ensures that computational resources are allocated efficiently, focusing on areas of the scene where finer sampling is most beneficial.

The adjustment of $\delta_i$ is based on a heuristic that considers several factors, including the viewer’s distance from rendered objects, the calculated gradient of viewer position, and framerate data collected in real time.

\subsection{QuiQ}

We train QuiQ through a multi-stage learning process \cite{Park19} designed to reduce training time by several orders of magnitude while maintaining a negligible loss in efficacy. First, a demo sequence is used to benchmark the computing limits of the target system. We found the benefits of increasing the duration of this data collection stage are minimal after 25 seconds. 

Next, the collected data are used to select a closely matching prerecorded dataset. This process allows the usage of training data that would otherwise take several minutes to collect. Specifically, this process uses the k-Nearest Neighbors (k-NN) algorithm to identify the most closely matching dataset. This larger dataset is used to train the reward model for the Q-learning resolution controller. In particular, we utilize ridge regression to construct a model designed to predict framerate from several rendering parameters. 

\begin{table}
  \caption{Ablation over dedicated QuiQ training time. Results are averaged over 3 scenes from the Tanks \& Temples dataset. We find only minimal improvement in resolution and FPS by increasing training time past 60 seconds.
}
  \begin{tabular}{lcccc}
    \toprule
    \ & PSNR & SSIM & LPIPS & \ FPS\\
    \midrule
    Ours-30s & 25.11 & 0.738 & 0.079 & \ 103\\
    Ours-60s & 27.52 & 0.773 & 0.068 & \ 104\\
    Ours-120s & 28.31 & 0.792 & 0.062 & \ 106\\
  \bottomrule
\end{tabular}
\end{table}

The Q-learning algorithm is trained on a sequence of predetermined inputs similar to the initial data collection stage and consequently rewarded by the ridge regression model. We created two implementations of the QuiQ training sequence – an asynchronous model that trains seamlessly in the background of the renderer and an independent model with results detailed in Table 1 that can fully prepare the QuiQ resolution controller in under a minute of dedicated processing time. These models are outlined in Fig. 2.

\section{Results}

We present our results on the VistaFlow framework through time and resolution comparisons with previous methods, showing comparable results in model optimization time and significantly improved performance in framerate stability on a predetermined rendering benchmark.

\begin{table}[H]
  \caption{Results compared to previous methods. We find that VistaFlow outperforms other rendering techniques by nearly a full order of magnitude in FPS while maintaining comparable (if not superior) resolution.
}
  \begin{tabular}{lcccc}
    \toprule
    \ & PSNR & SSIM & LPIPS & \ FPS\\
    \midrule
    NeRF & 31.69 & 0.953 & 0.068 & \ 0.045\\
    SRN & 22.26 & 0.846 & 0.17 & \ 0.909\\
    Neural Volumes & 26.05 & 0.893 & 0.16 & \ 3.33\\
    NSVF & 31.75 & 0.953 & 0.047 & \ 0.815\\
    AutoInt & 25.55 & 0.911 & 0.17 & \ 0.38\\
    PlenOctree & 31.76 & 0.958 & 0.053 & \ 6.97\\
    VistaFlow & 34.22 & 0.954 & 0.072 & \ 67.26\\
  \bottomrule
\end{tabular}
\end{table}

\begin{figure}[h]
  \centering
  \includegraphics[width=\linewidth]{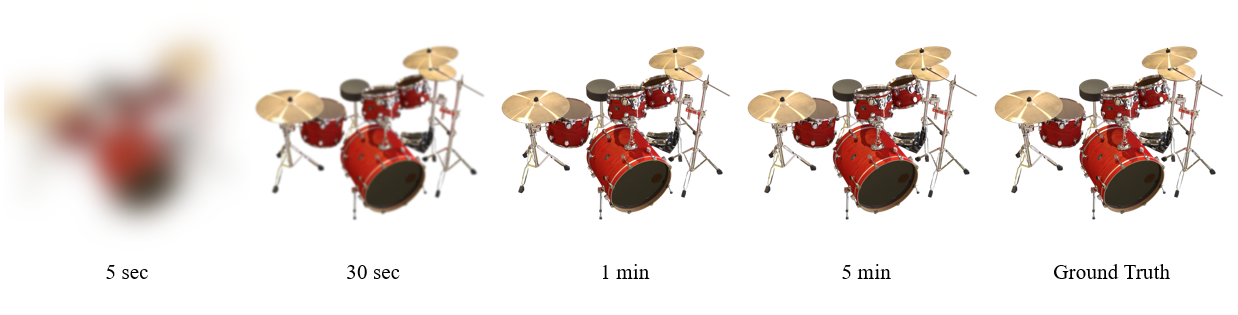}
  \caption{Results from our direct PlenOctree training method. We demonstrate efficient model optimization in minutes. Shown above is “drums” from the NeRF Synthetic dataset.
}
\end{figure}

\begin{figure}[h]
  \centering
  \includegraphics[width=\linewidth]{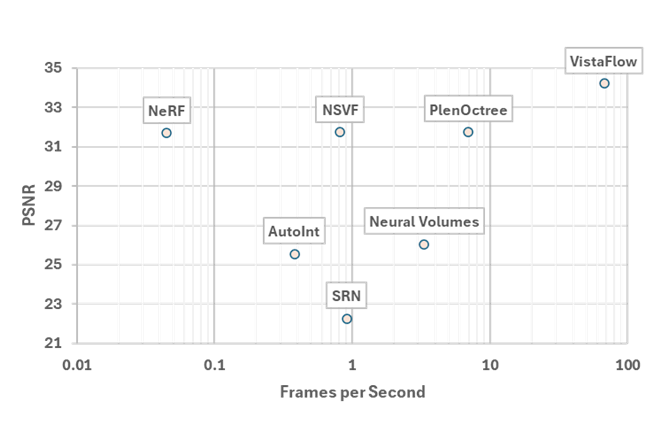}
  \caption{Quantitative results on NeRF Synthetic scenes. VistaFlow outperforms every previous method in both PSNR (which represents model accuracy) and FPS
}
\end{figure}

\begin{figure}[h]
  \centering
  \includegraphics[width=\linewidth]{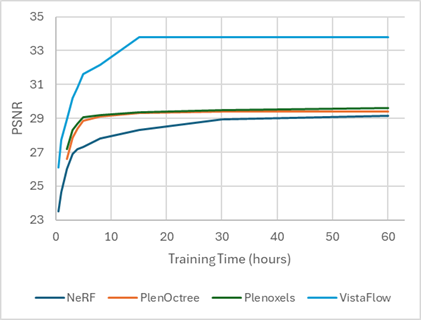}
  \caption{Training curves on the NeRF Synthetic drums scene. We find that VistaFlow trains both faster and more efficiently than previous methods. We maintain this significant gap in PSNR even as training time increases.
}
\end{figure}

\begin{figure}[h]
  \centering
  \includegraphics[width=\linewidth]{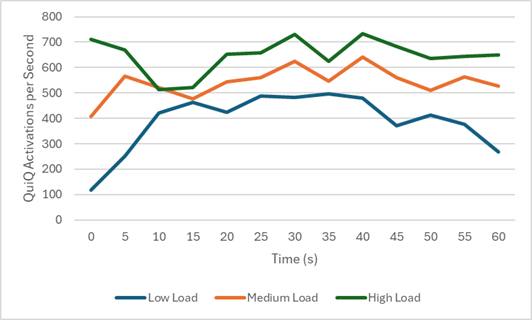}
  \caption{QuiQ Activations per second at varying levels of computational load as measured by CPU usage. QuiQ adjusts ray sampling parameters so rapidly that changes are invisible in the final output. This frequent activation is what allows QuiQ to achieve such impressive PSNR, SSIM, and LPIPS scores.
}
\end{figure}
\section{Conclusion}

The introduction of VistaFlow marks a significant leap forward in the domain of volumetric reconstruction, offering a scalable and dynamic solution to the challenges of photorealistic 3D scene rendering. By integrating dynamic resolution management through Q-learning, VistaFlow addresses the critical need for adaptable rendering quality in varied computational environments. This innovation not only enhances the feasibility of high-quality 3D rendering on less capable hardware but also propels the accessibility of volumetric reconstruction to a broader audience. The strategic employment of QuiQ for real-time resolution adjustment showcases a novel approach to maintaining high frame rates without compromising the integrity of the visual output. Moreover, the reliance on the PlenOctree data structure, devoid of Neural Radiance Fields, underscores a pivotal shift towards more efficient rendering techniques that eschew heavy computational demands.

\subsection{Limitations and Future Work}

While VistaFlow represents a significant advancement in volumetric reconstruction, a notable challenge is the occurrence of visual artifacts, particularly in highly detailed or dynamic scenes. This can detract from the overall realism and immersion of the rendered environment. These artifacts often arise due to the resolution management system struggling to adequately balance detail and performance across varying computational landscapes. Additionally, VistaFlow's current design primarily targets environments with balanced CPU and GPU capabilities; however, in CPU-bottlenecked workstations, the framework may not achieve its full potential due to a reduction in reliance on GPU calculation for VistaFlow compared to previous methods. Addressing this issue necessitates the development of an implementation that is better optimized for CPU-intensive scenarios, ensuring that VistaFlow can deliver consistent, high-quality rendering across a broader spectrum of hardware configurations. Future iterations will focus on minimizing visual artifacts through improved resolution management algorithms and enhancing CPU efficiency to broaden accessibility and applicability.

\begin{acks}
We thank the Illinois Mathematics and Science Academy ITS team for providing us with computing resources.
\end{acks}

\bibliographystyle{ACM-Reference-Format}
\bibliography{references}


\begin{thebibliography}{13}


\ifx \showCODEN    \undefined \def \showCODEN     #1{\unskip}     \fi
\ifx \showISBNx    \undefined \def \showISBNx     #1{\unskip}     \fi
\ifx \showISBNxiii \undefined \def \showISBNxiii  #1{\unskip}     \fi
\ifx \showISSN     \undefined \def \showISSN      #1{\unskip}     \fi
\ifx \showLCCN     \undefined \def \showLCCN      #1{\unskip}     \fi
\ifx \shownote     \undefined \def \shownote      #1{#1}          \fi
\ifx \showarticletitle \undefined \def \showarticletitle #1{#1}   \fi
\ifx \showURL      \undefined \def \showURL       {\relax}        \fi
\providecommand\bibfield[2]{#2}
\providecommand\bibinfo[2]{#2}
\providecommand\natexlab[1]{#1}
\providecommand\showeprint[2][]{arXiv:#2}

\bibitem[Aliev et~al\mbox{.}(2019)]%
        {Aliev19}
\bibfield{author}{\bibinfo{person}{K. Aliev}, \bibinfo{person}{A. Sevastopolsky}, \bibinfo{person}{M. Kolos}, \bibinfo{person}{D. Ulyanov}, {and} \bibinfo{person}{V. Lempitsky}.} \bibinfo{year}{2019}\natexlab{}.
\newblock \showarticletitle{Neural Point-Based Graphics}.
\newblock \bibinfo{journal}{\emph{arXiv (Cornell University)}} (\bibinfo{year}{2019}).
\newblock
\href{https://doi.org/10.48550/arxiv.1906.08240}{doi:\nolinkurl{10.48550/arxiv.1906.08240}}


\bibitem[Barron(2021)]%
        {Barron21}
\bibfield{author}{\bibinfo{person}{J.~T. Barron}.} \bibinfo{year}{2021}\natexlab{}.
\newblock \showarticletitle{MIP-NERF: A Multiscale Representation for Anti-Aliasing Neural Radiance Fields}.
\newblock \bibinfo{journal}{\emph{arXiv.org}} (\bibinfo{date}{March} \bibinfo{year}{2021}).
\newblock
\urldef\tempurl%
\url{https://arxiv.org/abs/2103.13415}
\showURL{%
\tempurl}


\bibitem[Chen et~al\mbox{.}(2022)]%
        {Chen22}
\bibfield{author}{\bibinfo{person}{Z. Chen}, \bibinfo{person}{T. Funkhouser}, \bibinfo{person}{P. Hedman}, {and} \bibinfo{person}{A. Tagliasacchi}.} \bibinfo{year}{2022}\natexlab{}.
\newblock \showarticletitle{MobileNERF: Exploiting the Polygon Rasterization Pipeline for Efficient Neural Field Rendering on Mobile Architectures}.
\newblock \bibinfo{journal}{\emph{arXiv (Cornell University)}} (\bibinfo{year}{2022}).
\newblock
\href{https://doi.org/10.48550/arxiv.2208.00277}{doi:\nolinkurl{10.48550/arxiv.2208.00277}}


\bibitem[Garbin et~al\mbox{.}(2021)]%
        {Garbin21}
\bibfield{author}{\bibinfo{person}{S.~J. Garbin}, \bibinfo{person}{M. Kowalski}, \bibinfo{person}{M.~S. Johnson}, \bibinfo{person}{J. Shotton}, {and} \bibinfo{person}{J. Valentin}.} \bibinfo{year}{2021}\natexlab{}.
\newblock \showarticletitle{FastNERF: High-Fidelity Neural Rendering at 200FPS}.
\newblock \bibinfo{journal}{\emph{arXiv (Cornell University)}} (\bibinfo{year}{2021}).
\newblock
\href{https://doi.org/10.48550/arxiv.2103.10380}{doi:\nolinkurl{10.48550/arxiv.2103.10380}}


\bibitem[Kerbl et~al\mbox{.}(2023)]%
        {Kerbl23}
\bibfield{author}{\bibinfo{person}{B. Kerbl}, \bibinfo{person}{G. Kopanas}, \bibinfo{person}{T. Leimkühler}, {and} \bibinfo{person}{G. Drettakis}.} \bibinfo{year}{2023}\natexlab{}.
\newblock \showarticletitle{3D Gaussian Splatting for Real-Time Radiance Field Rendering}.
\newblock \bibinfo{journal}{\emph{arXiv (Cornell University)}} (\bibinfo{year}{2023}).
\newblock
\href{https://doi.org/10.48550/arxiv.2308.04079}{doi:\nolinkurl{10.48550/arxiv.2308.04079}}


\bibitem[Knapitsch et~al\mbox{.}(2017)]%
        {Knapitsch17}
\bibfield{author}{\bibinfo{person}{A. Knapitsch}, \bibinfo{person}{J. Park}, \bibinfo{person}{Q. Zhou}, {and} \bibinfo{person}{V. Koltun}.} \bibinfo{year}{2017}\natexlab{}.
\newblock \showarticletitle{Tanks and Temples}.
\newblock \bibinfo{journal}{\emph{ACM Transactions on Graphics}} \bibinfo{volume}{36}, \bibinfo{number}{4} (\bibinfo{year}{2017}), \bibinfo{pages}{1--13}.
\newblock
\href{https://doi.org/10.1145/3072959.3073599}{doi:\nolinkurl{10.1145/3072959.3073599}}


\bibitem[Mildenhall et~al\mbox{.}(2020)]%
        {Mildenhall20}
\bibfield{author}{\bibinfo{person}{B. Mildenhall}, \bibinfo{person}{P.~P. Srinivasan}, \bibinfo{person}{M. Tancik}, \bibinfo{person}{J.~T. Barron}, \bibinfo{person}{R. Ramamoorthi}, {and} \bibinfo{person}{R. Ng}.} \bibinfo{year}{2020}\natexlab{}.
\newblock \showarticletitle{NeRF: Representing Scenes as Neural Radiance Fields for View Synthesis}.
\newblock \bibinfo{journal}{\emph{arXiv (Cornell University)}} (\bibinfo{year}{2020}).
\newblock
\href{https://doi.org/10.48550/arxiv.2003.08934}{doi:\nolinkurl{10.48550/arxiv.2003.08934}}


\bibitem[Park et~al\mbox{.}(2019)]%
        {Park19}
\bibfield{author}{\bibinfo{person}{J.~J. Park}, \bibinfo{person}{P. Florence}, \bibinfo{person}{J. Straub}, \bibinfo{person}{R. Newcombe}, {and} \bibinfo{person}{S. Lovegrove}.} \bibinfo{year}{2019}\natexlab{}.
\newblock \showarticletitle{DeepSDF: Learning Continuous Signed Distance Functions for Shape Representation}.
\newblock \bibinfo{journal}{\emph{arXiv (Cornell University)}} (\bibinfo{year}{2019}).
\newblock
\href{https://doi.org/10.48550/arxiv.1901.05103}{doi:\nolinkurl{10.48550/arxiv.1901.05103}}


\bibitem[Piala and Clark(2021)]%
        {Piala21}
\bibfield{author}{\bibinfo{person}{M. Piala} {and} \bibinfo{person}{R.~J. Clark}.} \bibinfo{year}{2021}\natexlab{}.
\newblock \showarticletitle{TERMiNERF: Ray Termination Prediction for Efficient Neural Rendering}.
\newblock \bibinfo{journal}{\emph{arXiv (Cornell University)}} (\bibinfo{year}{2021}).
\newblock
\href{https://doi.org/10.48550/arxiv.2111.03643}{doi:\nolinkurl{10.48550/arxiv.2111.03643}}


\bibitem[Reiser et~al\mbox{.}(2021)]%
        {Reiser21}
\bibfield{author}{\bibinfo{person}{C. Reiser}, \bibinfo{person}{S. Peng}, \bibinfo{person}{Y. Liao}, {and} \bibinfo{person}{A. Geiger}.} \bibinfo{year}{2021}\natexlab{}.
\newblock \showarticletitle{KiloNeRF: Speeding Up Neural Radiance Fields with Thousands of Tiny MLPs}.
\newblock \bibinfo{journal}{\emph{arXiv (Cornell University)}} (\bibinfo{year}{2021}).
\newblock
\href{https://doi.org/10.48550/arxiv.2103.13744}{doi:\nolinkurl{10.48550/arxiv.2103.13744}}


\bibitem[Seitz and Dyer(2002)]%
        {Seitz02}
\bibfield{author}{\bibinfo{person}{S.~M. Seitz} {and} \bibinfo{person}{C.~R. Dyer}.} \bibinfo{year}{2002}\natexlab{}.
\newblock \showarticletitle{Photorealistic Scene Reconstruction by Voxel Coloring}. In \bibinfo{booktitle}{\emph{Proceedings of IEEE Computer Society Conference on Computer Vision and Pattern Recognition}}.
\newblock
\href{https://doi.org/10.1109/cvpr.1997.609462}{doi:\nolinkurl{10.1109/cvpr.1997.609462}}


\bibitem[Yu et~al\mbox{.}(2021a)]%
        {Yu21plenoxels}
\bibfield{author}{\bibinfo{person}{A. Yu}, \bibinfo{person}{S. Fridovich-Keil}, \bibinfo{person}{M. Tancik}, \bibinfo{person}{Q. Chen}, \bibinfo{person}{B. Recht}, {and} \bibinfo{person}{A. Kanazawa}.} \bibinfo{year}{2021}\natexlab{a}.
\newblock \showarticletitle{Plenoxels: Radiance Fields Without Neural Networks}.
\newblock \bibinfo{journal}{\emph{arXiv (Cornell University)}} (\bibinfo{year}{2021}).
\newblock
\href{https://doi.org/10.48550/arxiv.2112.05131}{doi:\nolinkurl{10.48550/arxiv.2112.05131}}


\bibitem[Yu et~al\mbox{.}(2021b)]%
        {Yu21plenoctrees}
\bibfield{author}{\bibinfo{person}{A.~C.~L. Yu}, \bibinfo{person}{R. Li}, \bibinfo{person}{M. Tancik}, \bibinfo{person}{H. Li}, \bibinfo{person}{R. Ng}, {and} \bibinfo{person}{A. Kanazawa}.} \bibinfo{year}{2021}\natexlab{b}.
\newblock \showarticletitle{PlenOctrees for Real-Time Rendering of Neural Radiance Fields}.
\newblock \bibinfo{journal}{\emph{arXiv (Cornell University)}} (\bibinfo{year}{2021}).
\newblock
\href{https://doi.org/10.48550/arxiv.2103.14024}{doi:\nolinkurl{10.48550/arxiv.2103.14024}}


\end{thebibliography}

\end{document}